%% file: 0_main.tex
\documentclass[11pt,a4paper]{article}
\usepackage[hyperref]{emnlp2020}
\usepackage{times}
\usepackage{latexsym}
\usepackage{relsize}
\usepackage{tabularx, soul, enumitem,booktabs, multirow}

\usepackage[hang,flushmargin]{footmisc}
\usepackage{microtype}

\usepackage{graphicx}
\usepackage[T1]{fontenc}
\usepackage{topcapt}
\usepackage{booktabs}
\usepackage{enumitem}
\newtheorem{exmp}{Example}
\usepackage{amsmath}
\usepackage{amssymb}

\newcommand{\apphrase}[1]{{\small{#1}}}
\newcommand{\subjqa}{{\textsc{SubjQA}}}
\newcommand{\extraction}[2]{\guilsinglleft`\textit{#1}', `\textit{#2}'\guilsinglright}

\aclfinalcopy 
\def\aclpaperid{595} 

\title{\textsc{SubjQA}: A Dataset for Subjectivity and Review Comprehension}

\author{Johannes Bjerva$^{1,3}$\thanks{JB and NB contributed equally to this work.} \text{ }  Nikita Bhutani$^{2*}$ \text{ } \\ \textbf{Behzad Golshan}$^2$ \text{ } \textbf{Wang-Chiew Tan}$^2$ \text{ } \textbf{Isabelle Augenstein}$^1$ \\
${}^1$Department of Computer Science, University of Copenhagen \\
${}^2$Megagon Labs, Mountain View \\
${}^3$Department of Computer Science, Aalborg University \\
{\tt jbjerva@cs.aau.dk, augenstein@di.ku.dk}\\{\tt \{nikita,behzad,wangchiew\}@megagon.ai}
}

\date{}

\begin{document}
\maketitle

\begin{abstract}
\input{1_abstract}
\end{abstract}

\section{Introduction}
\input{2_introduction}

\section{Subjectivity}\label{sec:subjectivity}
\input{3_subjectivity}

\section{Data Collection}\label{sec:data_collection}
\input{6_collection}

\section{Dataset Analysis}\label{sec:analyses}
\input{7_dataset_analysis}

\input{4_model}

\input{5_experiments}\label{sec:experiments}

\section{Related Work}\label{sec:related_work}
\input{8_related}

\section{Conclusion}\label{sec:conclusion}

In this paper we investigate subjectivity in QA by leveraging end-to-end architectures.
We release \subjqa, a question-answering dataset which contains subjectivity labels for both questions and answers. 
The dataset allows i) evaluation and development of architectures for subjective content, and ii) investigation of subjectivity and its interactions in broad and diverse contexts.
We further implement a subjectivity-aware model and evaluate it, along with 4 strong baseline models.
We hope this dataset opens new avenues for research on querying subjective content, and into subjectivity in general.

\section*{Acknowledgements}
We are grateful to the Nordic Language Processing Laboratory (NLPL) for providing access to its supercluster infrastructure, and the anonymous reviewers for their helpful feedback.

\bibliography{subjqa,anthology}
\bibliographystyle{acl_natbib}

\clearpage
\appendix

\section{Appendices}
\input{appendix.tex}

\end{document}

%% file: 1_abstract.tex
Subjectivity is the expression of internal opinions or beliefs which cannot be objectively observed or verified, and has been shown to be important for sentiment analysis and word-sense disambiguation.
Furthermore, subjectivity is an important aspect of user-generated data.
In spite of this, subjectivity has not been investigated in contexts where such data is widespread, such as in question answering (QA). 
We develop a new dataset which allows us to investigate this relationship.
We find that subjectivity is an important feature in the case of QA, albeit with more intricate interactions between subjectivity and QA performance than found in previous work on sentiment analysis.
For instance, a subjective question may or may not be associated with a subjective answer.
We release an English QA dataset (\subjqa{}) based on customer reviews, containing subjectivity annotations for questions and answer spans across 6 domains. 

%% file: 2_introduction.tex
Subjectivity is ubiquitous in our use of language \citep{banfield1982unspeakable,quirk,wiebe-etal-1999-development,benamara2017evaluative}, and is therefore an important aspect to consider in Natural Language Processing (NLP).
For example, subjectivity can be associated with different senses of the same word. \textsc{boiling} is objective in the context of \textit{hot water}, but subjective in the context of a person \textit{boiling with anger} \citep{wiebe-mihalcea-2006-word}.
The same applies to sentences in discourse contexts \citep{pang2004sentimental}.
While early work has shown subjectivity to be an important feature for low-level tasks such as word-sense disambiguation and sentiment analysis, subjectivity in NLP has not been explored in many contexts where it is prevalent.

In recent years, there has been renewed interest in areas of NLP for which subjectivity is important, and a specific topic of interest is question answering (QA). This includes work on aspect extraction~\cite{poriaCG16}, opinion mining~\cite{sunLC17} and community QA~\cite{GuptaKCRL19}.
Many QA systems are based on representation learning architectures (e.g.~\citet{bert,gpt2}) that are typically trained on factual, encyclopedic knowledge such as Wikipedia or books. 
It is unclear if these architectures can handle subjective statements such as those that appear in reviews.


The interactions between QA and subjectivity are even more relevant today as users' natural search criteria in many domains, including products and services, have become increasingly subjective.
According to \citet{mcauley2016addressing}, around 20\% of product queries were labeled as being ‘subjective’ by workers.
Their questions can often be answered by online customer reviews, which tend to be highly subjective as well.
Although QA over customer reviews has gained traction recently with the availability of new datasets and architectures~\citep{Grail2019,GuptaKCRL19,FanFSLW19,huxu2019,li2019subjective}, these are agnostic with respect to how subjectivity is expressed in the questions and the reviews. Furthermore, the datasets are either very small (< 2000 questions) or have target-specific question types (e.g., yes-no). Most QA datasets and systems focus on answering questions over factual data such as Wikipedia articles and News \citep{JoshiCWZ17,TrischlerWYHSBS17,RajpurkarJL18,abdou-etal-2019-x,ReddyCM19}.
In this work, on the other hand, we focus on QA over subjective data from reviews on product and service websites.

In this work, we investigate the relation between subjectivity and question answering (QA) in the context of customer reviews.
As no such QA dataset exists, we construct \subjqa{}.\footnote{\subjqa{} is available at \url{https://github.com/megagonlabs/SubjQA}}
In order to capture subjectivity, our data collection method builds on  recent developments in opinion extraction and matrix factorisation, instead of relying on linguistic similarity between questions and reviews~\citep{GuptaKCRL19}. 
\subjqa{} includes over 10,000 English examples spanning 6 domains that cover both products and services. 
We find a large percentage of subjective questions and answers in \subjqa{}, as 73\% of the questions are subjective and 74\% of the answers are subjective.
Experiments show that existing QA systems trained to find factual answers struggle with subjective questions and reviews. For instance, fine-tuning \textsc{BERT} \citep{bert}, a state-of-the-art QA model, yields $92.9\%$ $F_1$ on SQuAD~\cite{RajpurkarZLL16}, but only achieves a mean score of $74.1\%$ $F_1$ across the domains in \subjqa{}.

To demonstrate the importance of subjectivity in QA, we develop a subjectivity-aware QA model that extends an existing QA model in a multi-task learning paradigm. It is trained to predict the subjectivity label and answer span simultaneously, and does not require subjectivity labels at test time.
This QA model achieves $76.3\%$ $F_1$ on an average over different domains of \subjqa{}.


\paragraph{Contributions} (i) We release a challenging QA dataset with subjectivity labels for questions and answers, spanning 6 domains; 
(ii) We investigate the relationship between subjectivity and QA;
(iii) We develop a subjectivity-aware QA model;
(iv) We verify the findings of previous work on subjectivity, using recent NLP architectures.


%% file: 3_subjectivity.tex
Written text, as an expression of language, contains information on several linguistic levels, many of which have been thoroughly explored in NLP.\footnote{Subjectivity is not restricted to written texts, although we focus on this modality here.}
For instance, both the semantic content of text and the (surface) forms of words and sentences, as expressed through syntax and morphology, have been at the core of the field for decades.
However, another level of information can be found when trying to observe or encode the so-called \textit{private states} of the writer \citep{quirk}. 
Examples of private states include the opinions and beliefs of a writer, and can concretely be said to not be available for verification or objective observation. 
It is this type of state which is referred to as \textit{subjectivity} \citep{banfield1982unspeakable,banea2011multilingual}.

Whereas subjectivity has been investigated in isolation, it can be argued that subjectivity is only meaningful given sufficient context.
Regardless, much previous work has focused on annotating words \citep{heise2001project}, word senses \citep{durkin1989polysemy,wiebe-mihalcea-2006-word,akkaya2009subjectivity}, or sentences \citep{pang2004sentimental}, with notable exceptions such as \citet{wiebe2005annotating,banea2010multilingual}, who investigate subjectivity in phrases in the context of a text or conversation. There is limited work that studies subjectivity using architectures that allow for contexts to be incorporated efficiently \citep{vaswani2017attention}.

As subjectivity relies heavily on context, and we have access to methods which can encode such context, what then of access to data which encodes subjectivity?
We argue that in order to fully investigate research questions dealing with subjectivity in contexts, a large-scale dataset is needed.
We choose to frame this as a QA dataset, as it not only offers the potential to investigate interactions in a single contiguous document, but also allows interactions \textit{between} contexts, where parts may be subjective and other parts may be objective.
Concretely, one might seek to investigate the interactions between an objective question and a subjective answer, or vice-versa.


%% file: 6_collection.tex

\begin{figure}[tb]
    \centering
    \includegraphics[scale=0.45]{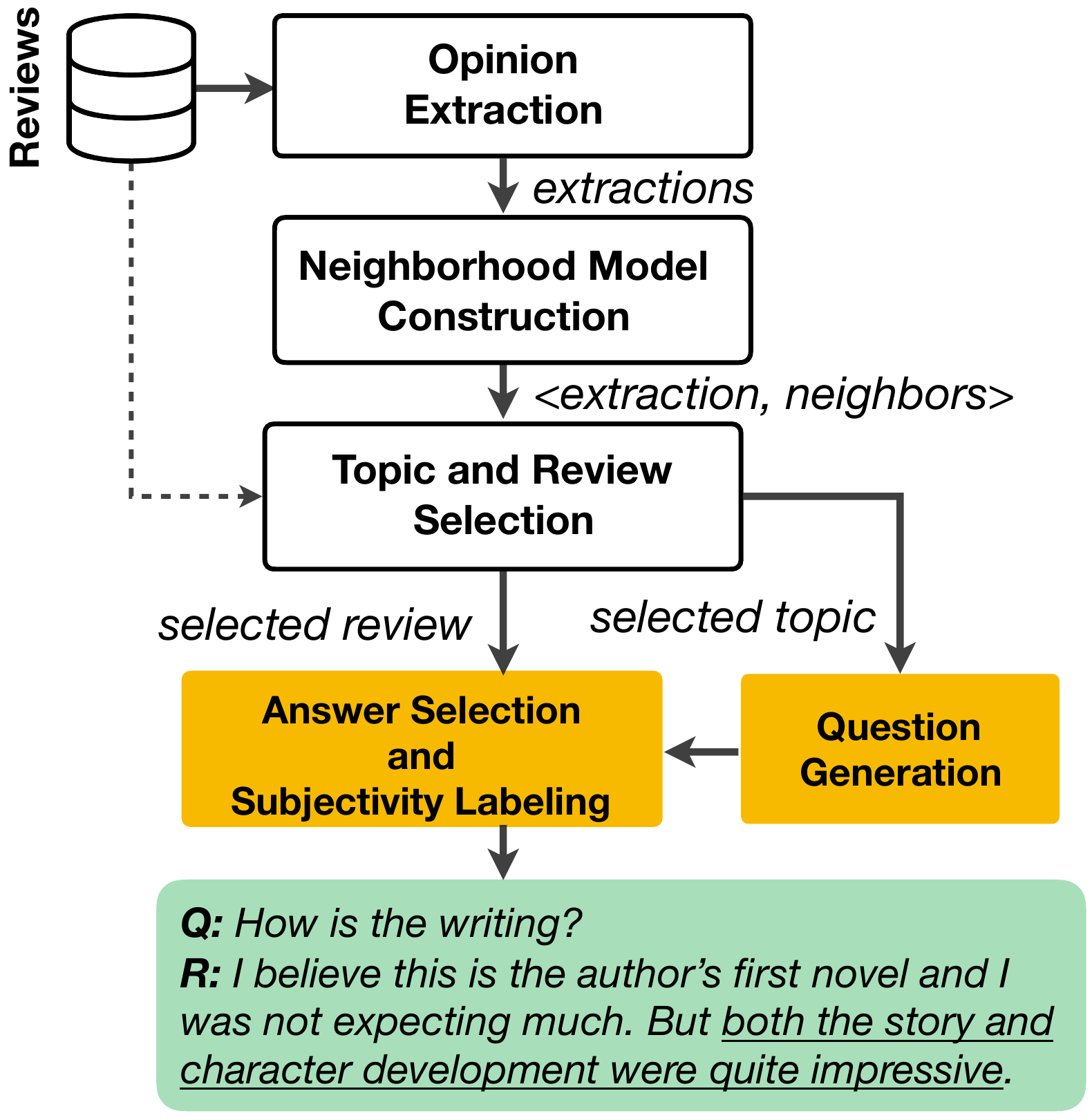}
    \caption{Our data collection pipeline}
    \label{fig:pipeline}
\end{figure}


We found two limitations of existing datasets and collection strategies that motivated us to create a new QA dataset to understand subjectivity in QA. First, data collection methods~\cite{GuptaKCRL19,huxu2019} often rely on the linguistic similarity between the questions and the reviews (e.g. information retrieval). However, subjective questions may not always use the same words or phrases as the review. 
Consider the examples below. 
The answer span `\emph{vegan dishes}' is similar to the question Q$_1$. 
The answer to the more subjective question Q$_2$, however, is linguistically quite different.

\begin{exmp}\label{ex:subj_example}
\normalfont{Q$_1$: Is the restaurant vegan friendly?}

\noindent
\normalfont{Review: [...] many \textbf{vegan dishes} on its menu.}

\smallskip
\noindent
\normalfont{Q$_2$: Does the restaurant have a romantic vibe?}

\noindent
\normalfont{Review: Amazing selection of wines, \textbf{perfect for a date night}.}
\end{exmp}

Secondly, existing review-based datasets are small and not very diverse in terms of question topics and types~\cite{xu2019bert,GuptaKCRL19}. We, therefore, consider reviews about both products and services from 6 different domains, namely TripAdvisor, Restaurants, Movies, Books, Electronics and Grocery. We use the data of~\citet{TripAdvisor} for TripAdvisor, and Yelp\footnote{\url{https://www.yelp.com/dataset}} data for Restaurants. We use the subsets for which an open-source opinion extractor was available~\citep{li2019subjective}. We use the data of~\citet{amazonreviews} that contains reviews from product pages of Amazon.com spanning multiple categories. We target categories that had more opinion expressions than others, determined by an opinion extractor.

Figure~\ref{fig:pipeline} depicts our data collection pipeline which builds upon recent developments in opinion extraction and matrix factorization. An opinion extractor is crucial to identify subjective or opinionated expressions, which other IR-based methods cannot. On the other hand, matrix factorization helps identify which of these expressions are related based on their co-occurrence in the review corpora, instead of their linguistic similarities. 
Relying on opinions instead of factual information to construct the dataset makes it inherently subjective. 
Furthermore, pairing questions and reviews based on related opinion expressions provides us some control over the diversity in the dataset.
To the best of our knowledge, we are the first to explore such a method to construct a subjective QA dataset.

Given a review corpus, we extract opinions about
various aspects of the items being reviewed (\emph{Opinion Extraction}). 
Consider the following review snippets and extractions.

\begin{exmp}\label{ex:pipeline_example}
\noindent
\normalfont{Review: [...] character development was quite impressive.}

\noindent
\normalfont{$e_1$:\extraction{impressive}{character development}}

\smallskip
\noindent
\normalfont{Review: 3 stars for good power and good writing.}

\noindent
\normalfont{$e_2$:\extraction{good}{writing}}

\end{exmp}

In the next (\emph{Neighborhood Model Construction}) step, we characterize the items being reviewed and their subjective extractions using latent features between two items. In particular, we use matrix factorization techniques~\cite{riedel2013relation}
to construct a neighborhood model $N$ via a set of weights $w_{e,e'}$, where each corresponds to a directed association strength between extraction $e$ and $e'$.
For instance, $e_1$ and $e_2$ in Example~\ref{ex:pipeline_example} could have a similarity score $0.93$.  This neighborhood model forms the core of data collection. We select a subset of extractions from $N$ as \emph{topics} (\emph{Topic Selection}) and ask crowd workers to translate them to natural language questions (\emph{Question Generation}). For each topic, a subset of its neighbors from $N$ and reviews which mention them are selected (\emph{Review Selection}). In this manner, question-review pairs are generated based on the neighborhood model.


Finally, we present each question-review pair to crowdworkers who highlight an answer span in the review. Additionally, they provide subjectivity scores for both the questions and the answer span. 



\subsection{Opinion Extraction}

An opinion extractor processes all reviews and finds extractions 
\guilsinglleft X,Y\guilsinglright\ where X represents an
opinion expressed on aspect Y.  Table~\ref{tab:extraction} shows sample extractions from different domains.
We use OpineDB~\cite{li2019subjective}, a state-of-the-art opinion extractor, for restaurants and hotels. For other domains where OpineDB was not available, we use the syntactic extraction patterns of ~\citet{abbasi2013aspect}.

\begin{table}
  \centering
    \resizebox{0.8\columnwidth}{!}{
    \begin{tabular}{lcc}
  \toprule  
  \textbf{Domain} & \textbf{Opinion Span} & \textbf{Aspect Span}  \\
  \midrule
  Restaurants & huge & lineup\\
  Hospitality & no free & wifi\\
  Books & hilarious & book\\
  Movies & not believable & characters\\
  Electronics & impressive & sound\\
  Grocery & high & sodium level\\
  \bottomrule
  \end{tabular}
  }
  \caption{Example extractions from different domains}
  \label{tab:extraction}
\end{table}  




\subsection{Neighborhood Model Construction}

\begin{figure}
    \centering
    \includegraphics[scale=0.30]{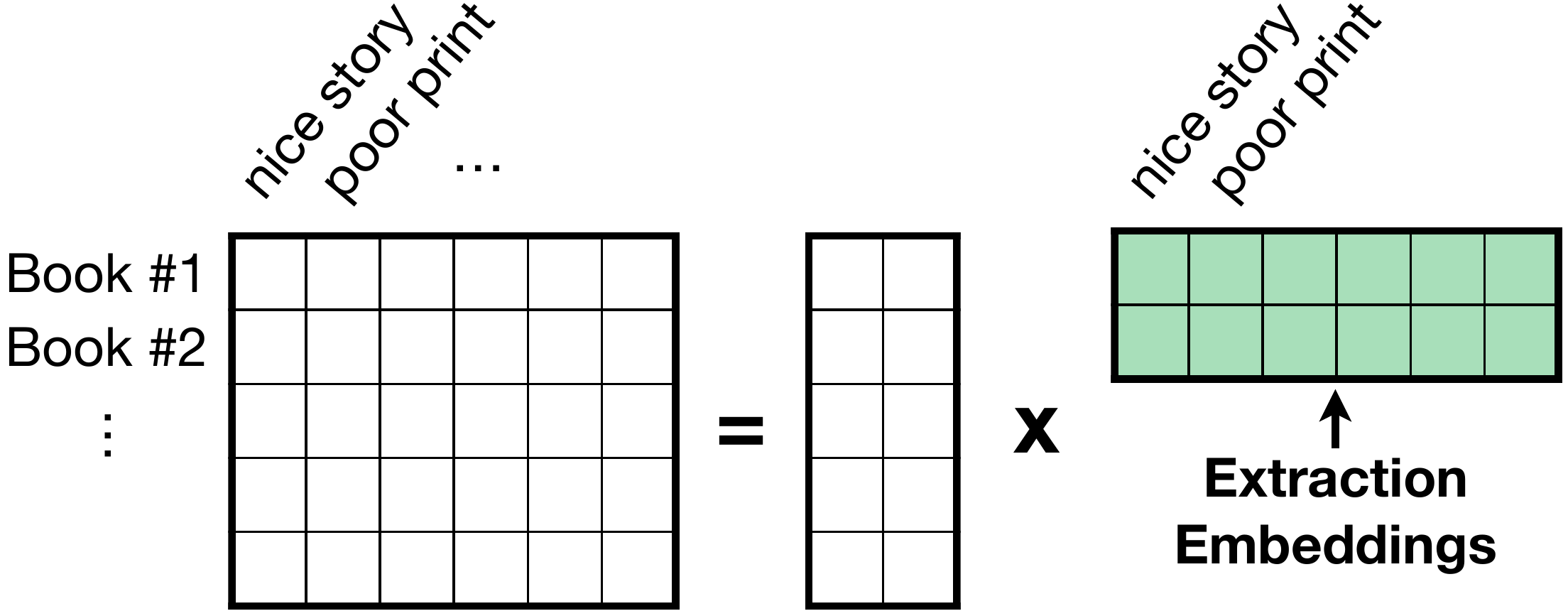}
    \caption{Learning representations of extractions via non-negative matrix factorization}
    \label{fig:nmf}
\end{figure}

We rely on matrix factorization to learn dense representations for items and extractions, and identify similar extractions. Such a method has been shown to be effective in building knowledge bases~\cite{sampo}. As depicted in Figure~\ref{fig:nmf}, we organize the extractions into a matrix $M$ where each row $i$ corresponds to an item being reviewed and
each column $j$ corresponds to an extraction. The value $M_{ij}$ denotes the frequency of extraction $e_j$ in reviews of item $x_i$. Given $M$ and a latent feature model $F$, we obtain extraction embeddings using non-negative matrix factorization.
Concretely, each value $M_{ij}$ is obtained from the dot product of two extractions of size $K^F$:
\begin{align}
    M^F_{ij} = \sum_{k}^{K^F} x_{i,k} e_{j,k}.
    \label{eq:matrixfac}
\end{align}
For each extraction, we find its \emph{neighbors} based on the cosine similarity of their embeddings.\footnote{Details about hyper-parameters are included in the Appendix.}


\subsection{Topic and Review Selection}
\label{subsec:queryselection}

We next identify a subset of extractions to be used as \emph{topics} for the questions. In order to maximize the diversity and difficulty in the dataset, we use the following criteria developed iteratively based on manual inspection followed by user experiments.


\begin{enumerate}[leftmargin=*,noitemsep,topsep=0pt]
    \item \textbf{Cosine Similarity:} We prune neighbors of an extraction which have low cosine similarity (< $0.8$). Irrelevant neighbors can lead to noisy topic-review pairs which would be marked \emph{non-answerable} by the annotators.
    
    \item \textbf{Semantic Similarity:} We prune neighbors that are linguistically similar (> $0.975$ similarity)\footnote{Using GloVe embeddings provided by spaCy.} as they yield easy topic-review pairs. 
    
    \item \textbf{Diversity}: To promote diversity in topics and reviews, we select extractions which have many ( $> 5$) neighbors.
    \item \textbf{Frequency}: To ensure selected topics are also popular, we select a topic if: a) its frequency is higher than the median frequency of all
    extractions, and b) it has at least one neighbor that is more frequent than the topic itself.
    
\end{enumerate}
We pair each topic with reviews that mention one of its neighbors. The key benefit of a factorization-based method is that it is not only based on linguistic similarity, and forces a QA system to understand subjectivity in questions and reviews.

\subsection{Question Generation}
Each selected topic is presented to a human annotator together with a review that mentions that topic. We ask the annotator to write a question about the topic that can be answered by the review. For example, \extraction{good}{writing} could be translated to ``Is the writing any good?" or ``How is the writing?". 


\subsection{Answer-Span and Subjectivity Labeling}

Lastly, we present each question and its corresponding review to human annotators (crowdworkers), who provides a subjectivity score to the question on a 1 to 5 scale based on whether it seeks an opinion (e.g., ``How good is this book?") or factual information (e.g., ``is this a hard-cover?"). Additionally, we ask them to highlight the shortest answer span in the review or mark the question as unanswerable. They also provide subjectivity scores for the answer spans. We provide details of our neighborhood model construction and crowdsourcing experiments in the Appendix.


%% file: 7_dataset_analysis.tex
In this section, we analyze the questions and answers to understand the properties of our \subjqa{} dataset. We present the dataset statistics in Section~\ref{subsec:data_stats}. We then analyze the diversity and difficulty of the questions. We also discuss the distributions of subjectivity and answerability in our dataset. Additionally, we manually inspect 100 randomly chosen questions from the development set in Section~\ref{subsec:data_quality} to understand the challenges posed by subjectivity of the questions and/or the answers.

\begin{table}
\centering
\resizebox{0.75\columnwidth}{!}{
	\begin{tabular}{lrrrr}
	\toprule
	\textbf{Domain} & \textbf{Train} & \textbf{Dev} & \textbf{Test} & \textbf{Total} \\
        \midrule
        TripAdvisor & 1165 & 230 & 512 & 1686 \\ 
        Restaurants & 1400 & 267 & 266 & 1683\\
        Movies & 1369 & 261 &  291 & 1677 \\
        Books & 1314 & 256 & 345 & 1668\\
        Electronics & 1295 & 255 & 358 & 1659 \\
        Grocery & 1124 & 218 & 591 & 1725 \\
        \bottomrule
	\end{tabular}
	}
  	\caption{No. of examples in each domain split.}
  	\label{tab:dataset_sizes}
\end{table}

\subsection{Data Statistics}\label{subsec:data_stats}

Table~\ref{tab:dataset_sizes} summarizes the number of examples we collected for different domains. To generate the train, development, and test splits, we partition the topics into training (80\%), dev (10\%) and test (10\%) sets. We partition the questions and reviews based on the partitioning of the topics.

\begin{table}
\fontsize{9}{11}\selectfont
\centering
\resizebox{\columnwidth}{!}{
	\begin{tabular}{lrrrr}
	\toprule
	\textbf{Domain} & \textbf{Review len} & \textbf{Q len}  & \textbf{A len} & \textbf{\% answerable}\\
        \midrule
        TripAdvisor & 187.25 & 5.66 &  6.71 & 78.17\\ 
        Restaurants & 185.40 & 5.44 & 6.67 & 60.72\\
        Movies & 331.56 & 5.59 & 7.32 & 55.69\\
        Books & 285.47 & 5.78 & 7.78 & 52.99\\
        Electronics & 249.44 & 5.56 & 6.98 & 58.89\\
        Grocery & 164.75 & 5.44 & 7.25 & 64.69\\
        \bottomrule
	\end{tabular}
	}
  	\caption{Domain statistics. Len denotes n tokens.}
  	\label{tab:data_len_stats}
\end{table}

\subsection{Difficulty and Diversity of Questions}

\begin{table}
\centering

\resizebox{\columnwidth}{!}{
	\begin{tabular}{lrrr}
	\toprule
	\textbf{Domain} & \textbf{\# questions} & \textbf{\# aspects} & \% \textbf{boolean Q}\\
        \midrule
        TripAdvisor & 1411 & 171 & 16.13\\ 
        Restaurants & 1553 & 238 & 17.29\\
        Movies & 1556 & 228 & 15.56\\
        Books & 1517 & 231 & 16.90\\
        Electronics & 1535 & 314 & 14.94\\
        Grocery & 1333 & 163 & 14.78\\
        \bottomrule
	\end{tabular}
	}
  	\caption{Diversity of questions and topics}
  	\label{tab:question_div}
\end{table}

\begin{figure}[t]
    \centering
    \includegraphics[width=0.9\columnwidth]{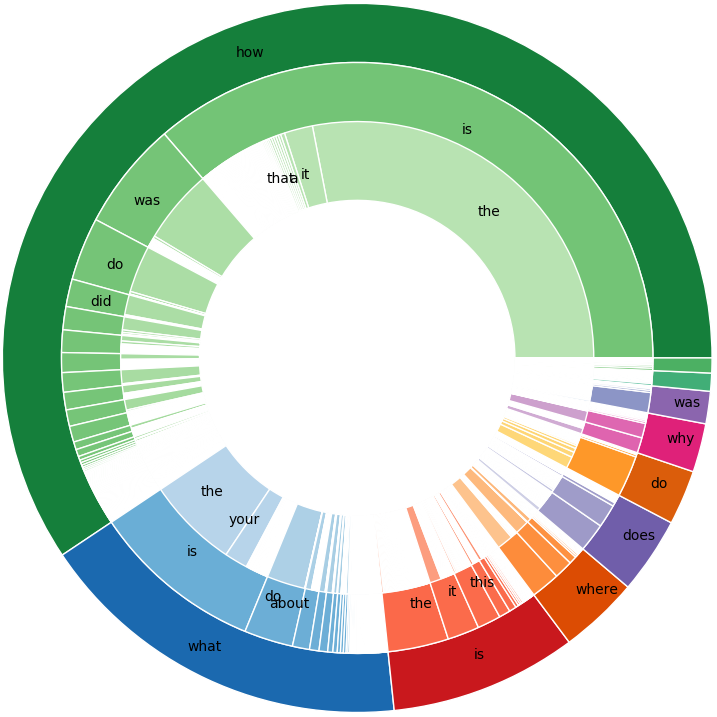}
    \caption{The distribution of prefixes of questions. The outermost ring shows unigram prefixes (e.g., 57.9\% questions start with \textit{how}). The middle and innermost rings correspond to bigrams and trigrams, respectively.}
    \label{fig:ques_trigram_prefix}
\end{figure}

\begin{table*}[ht]
\fontsize{9}{10}\selectfont
\centering
\setlength{\tabcolsep}{0.5em}
	\begin{tabularx}{0.9\textwidth}{llX}
	\toprule
	\textbf{Reasoning} & \textbf{Percent.} & \textbf{Example}\\
        \midrule
        Lexical & 18\% & {Q: How small was the hotel bathroom? \newline
                R: ...\textbf{Bathroom on the small} side with older fixtures...} \\ 
        \midrule
        Paraphrase & 28\% & {Q: How amazing was the end? \newline
                R: ...The \textbf{ending was absolutely awesome}, it makes the experience not so ... } \\
        \midrule
        Indirect & 43\% & {Q: How was the plot of the movie? \newline
                R: ...simply because there's so much going on, \textbf{so much action}, so many complex ..} \\ 
        \midrule
        N/A  & 11\% & {Q: How do you like the episode? \newline
                R: \textbf{For a show that I think was broadcast in HighDef}, it seems impossible that the...}\\ 
        \bottomrule
	\end{tabularx}
  	\caption{Types of reasoning required for the various domains. N/A indicates noisy or incorrect spans.}
  	\label{tab:dataset_reasoning}
\end{table*}

Table~\ref{tab:data_len_stats} shows that reviews in different domains tend to vary in length. Answer spans tend to be 6-7 tokens long, compared to 2-3 tokens in SQuAD.  

Table~\ref{tab:question_div} shows the number of distinct questions and topics in each domain. On average we collected 1500 questions covering 225 aspects. We also automatically categorize the boolean questions based on a lexicon of question prefixes.
Unlike other review-based QA datasets~\cite{GuptaKCRL19}, \subjqa{} contains more diverse questions, the majority of which are not yes/no questions.
The questions in \subjqa{} are also linguistically varied, as indicated by the trigram prefixes of the questions (Figure~\ref{fig:ques_trigram_prefix}). Most of the frequent trigram prefixes in \subjqa{} (e.g., \textit{how is the}, \textit{how was the}, \textit{how do you}) are almost missing in SQuAD and ~\citet{GuptaKCRL19}. The diversity of questions in \subjqa{} demonstrate challenges unique to the dataset.

\subsection{Data Quality Assessment}\label{subsec:data_quality}

We randomly sample 100 answerable questions to manually categorize them according to their reasoning types. Table~\ref{tab:dataset_reasoning} shows the distribution of the reasoning types and representative examples. As expected, since a large fraction of the questions are subjective, they cannot be simply answered using a keyword-search over the reviews or by paraphrasing the input question. Answering such questions requires a much deeper understanding of the reviews. Since the labels are crowdsourced, a small fraction of the answer spans are noisy. 

We also categorized the answers based on answer-types. We observed that 64\% of the answer spans were independent clauses (e.g., \textit{the staff was very helpful and friendly}), 25\% were noun phrases (e.g., \textit{great bed}) and 11\% were incomplete clauses/spans (e.g., \textit{so much action}). This supports our argument that subjective questions often cannot be answered simply by an adjective or noun phrase.

We rely on an automatically constructed opinion KB and collect labels for answer spans and subjectivity for all the question-review pairs.
As crowdworkers would label any spurious question-review pair as `unanswerable', this might increase the number of negative examples in the dataset. However, such examples are both much more difficult and more valuable than randomly paired questions and reviews. 
We found that the KBs used in our dataset collection achieved 35-50\% mean precision on various domains. 
We also found that approximately 48\% of a random set of 50 question-review pairs from SubjQA were marked unanswerable because of the unrelated opinion pairs in the KB.

\subsection{Answerability and Subjectivity}

The dataset construction relies on a neighborhood model generated automatically using factorization. It captures co-occurrence signals instead of linguistic signals. Consequently, the dataset generated is not guaranteed to only contain answerable questions. As expected, about 65\% of the questions in the dataset are answerable from the reviews (see Table~\ref{tab:data_subj}). However, unlike~\citet{GuptaKCRL19}, we do not predict answerability using a classifier. The answerability labels are provided by the crowdworkers instead, and are therefore more reliable.

Table~\ref{tab:data_subj} shows the subjectivity distribution in questions and answer spans across different domains. A vast majority of the questions we collected are subjective, which is not surprising since we selected topics from opinion extractions. A large fraction of the subjective questions ($\sim$70\%) were also answerable from their reviews. 


We also compare the subjectivity of questions with the subjectivity of answers. As can be seen in Table~\ref{tab:subjectivity_stats}, the subjectivity of an answer is strongly correlated with the subjectivity of the question. Subjective questions often have answers that are also subjective. Similarly, factual questions, with few exceptions, have factual answers. This indicates that a QA system must understand how subjectivity is expressed in a question to correctly find its answer. Most domains have 75\% subjective questions on average. However, the BERT-QA model fine-tuned on each domain achieves 80\% F1 on subjective questions in movies and books, but only achieves 67-73\% F1 on subjective questions in grocery and electronics. Future QA systems for user-generated content, such as for customer support, 
should therefore model subjectivity explicitly. 

\begin{table}
\centering
\resizebox{0.6\columnwidth}{!}{
	\begin{tabular}{lrr}
	\toprule
	& \textbf{subj. Q} & \textbf{fact. Q} \\
        \midrule
        subj. A & 79.8\% & 1.31\%  \\ 
        fact. A & 1.29\% & 17.58\% \\
        \bottomrule
	\end{tabular}
	}
  	\caption{Subjectivity distribution in \subjqa{}.}
  	\label{tab:subjectivity_stats}
\end{table}

\begin{table}
\centering
\resizebox{\columnwidth}{!}{
	\begin{tabular}{lrrr}
	\toprule
	\textbf{Domain} & \textbf{\% subj. Q} & \textbf{\% answerable}  & \textbf{\% subj. A}\\
        \midrule
        TripAdvisor & 74.49 & 83.20 & 75.20\\ 
        Restaurants & 76.11 & 65.72 & 76.29\\
        Movies & 74.41 & 62.09 & 74.59\\
        Books & 75.77 & 58.86 & 75.35 \\
        Electronics & 69.80 &  65.37 & 69.98\\
        Grocery & 73.21 & 70.22 & 73.15\\
        \bottomrule
	\end{tabular}
	}
  	\caption{Statistics on subjective Q, answerability, and subjective A per domain in \subjqa{}.}
  	\label{tab:data_subj}
\end{table}

\subsection{Comparison with other datasets}

Although this is not the first work to design a review-based QA dataset, no prior QA dataset targets the understanding of subjectivity in reviews.
AmazonQA \citep{GuptaKCRL19} is one of the largest review-based QA dataset, but differs from SubjQA in three core ways: 
i) they lack subjectivity labels;
ii) they are constructed using retrieval methods; and 
iii) they do not highlight answer spans in the reviews and instead provide an answerability label for the question given a set of reviews. 
These differences make it difficult for a fair comparison with their dataset. 
Some of the distinguishing characteristics of the two datasets include: 
(a) AmazonQA reports 61\% of the questions are answerable with 72\% precision (based on a classifier). In comparison, 65\% of the questions in SubjQA are answerable (based on human labels); 
(b) We provide opinion annotations and subjectivity labels for the questions, making it easier for researchers to target specific opinions and subjective text; and 
(c) 58\% of questions in SubjQA start with `how'. 
In comparison, most of the question prefixes (e.g., `will this work', `does it come with' etc.) in AmazonQA indicate product queries rather than opinion-based queries.





%% file: 4_model.tex
\section{Subjectivity Modelling}

We now turn to experiments on subjectivity, first investigating claims made by previous work, and whether they still hold when using recently developed architectures, before investigating how to model subjectivity in QA.

\subsection{Subjectivity in Sentiment Analysis}
\citet{pang2004sentimental} have shown that subjectivity is an important feature for sentiment analysis.
Sorting sentences by their estimated subjectivity scores, and only using the top $n$ such sentences, allows for a more efficient and better-performing sentiment analysis system, than when considering both subjective and objective sentences equally. 
We first investigate whether the same findings hold true when subjectivity is estimated using transformer-based architectures.\footnote{Using the BERT-base uncased model.} 
We predict the subjectivity of a sentence by passing its representation through a feed-forward neural network. 
We compare this with using subjectivity scores of \textsc{TextBlob}\footnote{\url{https://textblob.readthedocs.io/}}, a sentiment lexicon-based method, as a baseline. 


We evaluate this setup on subjectivity data from \citet{pang2004sentimental}\footnote{\url{http://www.cs.cornell.edu/people/pabo/movie-review-data/}} and the subjectivity labels made available in \subjqa{}.
Unsurprisingly, a contextually-aware classifier vastly outperforms a word-based classifier, highlighting the importance of context in subjectivity analysis (see Table~\ref{tab:clf}). 
Interestingly, however, predicting subjectivity in \subjqa{} is more challenging than in IMDB - we hypothesise that this is because \subjqa{} spans multiple domains.

\begin{table}[tb]
    \centering
   \resizebox{0.9\columnwidth}{!}{
    \begin{tabular}{lrr}
    \toprule
                   & \textbf{IMDB} & \textbf{\subjqa{}}  \\
    \midrule
        Word-based (\textsc{TextBlob})       & 61.90 &  57.50 \\ 
        BERT fine-tuned         & 88.20  & 62.77 \\
    \bottomrule
    \end{tabular}
    }
    \caption{Subjectivity prediction accuracies on IMDB data \citep{pang2004sentimental} and our dataset (\subjqa{}).}
    \label{tab:clf}
\end{table}

\begin{figure}
    \centering
    \includegraphics[width=0.9\columnwidth]{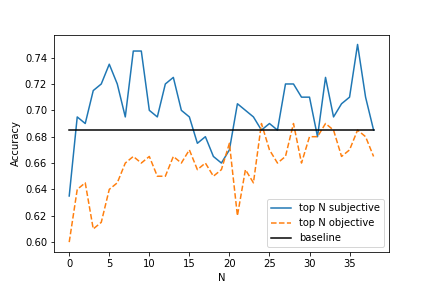}
    \caption{Sentiment Analysis accuracy using top N subj. sentences (blue), top N fact. sentences (orange dashed), compared to the all sentences baseline (black).} 
    \label{fig:sentiment_analysis}
\end{figure}

We further investigate if our subjectivity classifier is beneficial for sentiment analysis.
We replicate \citet{pang2004sentimental}, and perform sentiment analysis using the top $N$ subjective and objective sentences based on our system.
Figure~\ref{fig:sentiment_analysis} shows that giving a contextually-aware sentiment analysis model access to $N$ subjective sentences improves performance, as compared to using all sentences, or using $N$ objective sentences.


\subsection{Subjectivity-Aware QA Model}\label{sec:subj_model}

Given the importance of subjectivity in other NLP tasks, we investigate whether it is also an important feature for QA in \subjqa{}, with a relatively simple model.
We approach this by implementing a subjectivity-aware QA model, as an extension of our baseline models FastQA \citep{fastqa} in a multitask learning (MTL) paradigm \citep{caruana:1997}, as this approach has been shown to be useful for learning cross-task representations \citep{bjerva:2017,bjerva-2017-will,bingel-sogaard-2017-identifying,de-lhoneux-etal-2018-parameter,N18-1172,conf/emnlp2019/Augenstein,ruder2019latent}.
One concrete advantage of using MTL is that we do not need to have access to subjectivity labels at test time, as would be the case if we required subjectivity labels as a feature for each answer span.
Each input paragraph is encoded with a bidirectional LSTM \citep{lstm} over a sequence of word embeddings and contextual features ($\boldsymbol{\tilde{X}}$).
This encoding, $\boldsymbol{H^\prime}$, is passed through a hidden layer and a non-linearity:
\begin{align}
    \boldsymbol{H^\prime} &= \operatorname{Bi-LSTM}(\boldsymbol{\tilde{X}}) \\ 
    \boldsymbol{H} &= \tanh(B \boldsymbol{H^\prime}^\top) 
    \label{eq:fastqa}
\end{align}
We extend this implementation by adding two hidden layers of task-specific parameters ($\boldsymbol{W^n}$) associated with a second learning objective:
\begin{align}
    \boldsymbol{S^\prime} &= \operatorname{ReLU}(W^1\boldsymbol{H}) \\
    \boldsymbol{S} &= \operatorname{softmax(W^2\boldsymbol{S^\prime})}
    \label{eq:fastqa2}
\end{align}
\noindent In training, we randomly sample between the two tasks (QA and Subjectivity classification).

%% file: 5_experiments.tex
\begin{figure}
    \centering
    \includegraphics[scale=0.35]{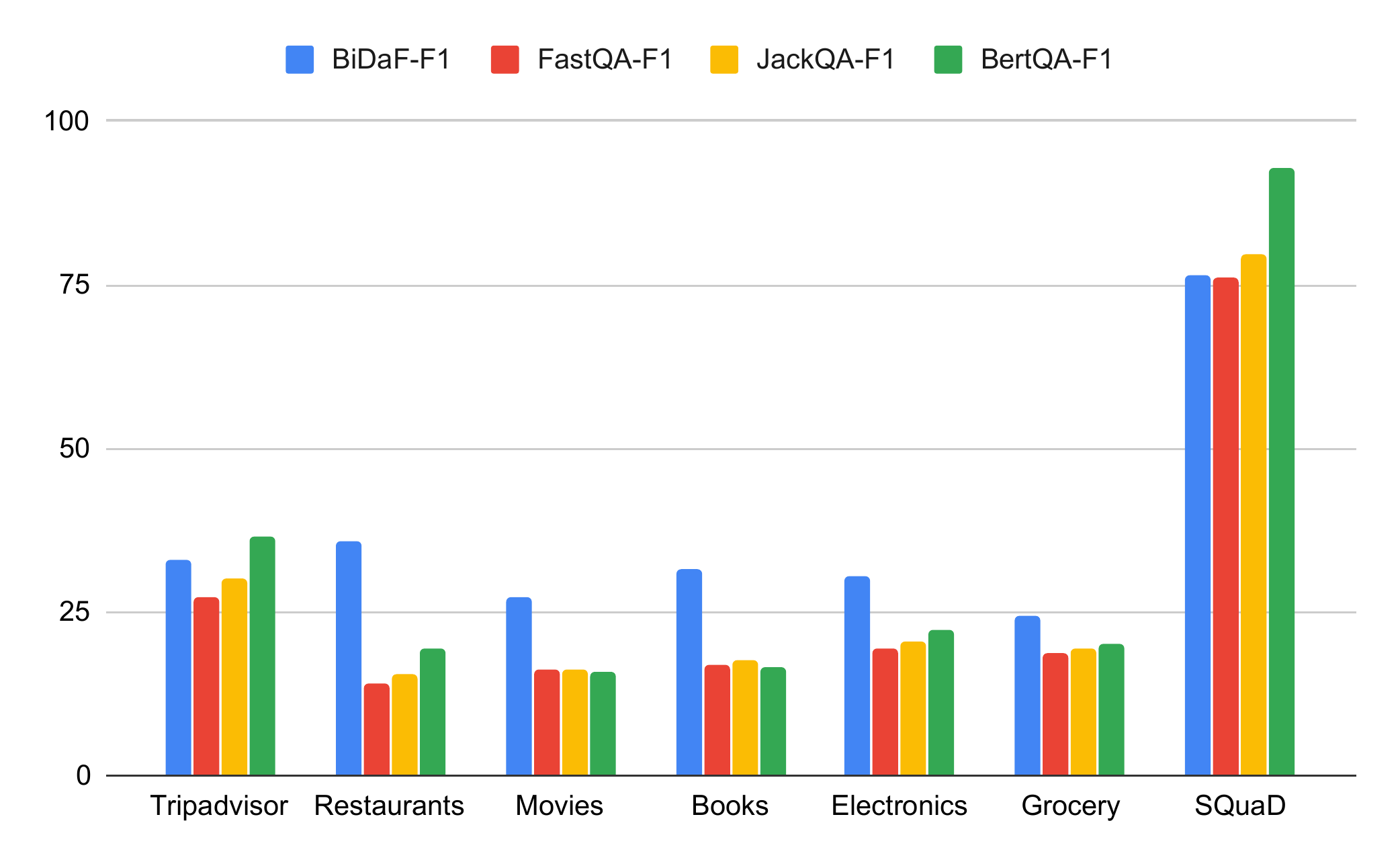}
    \caption{F1 scores of pre-trained out-of-the-box models on different domains in \subjqa{}. }
    \label{fig:qa_baselines}
\end{figure}

\begin{table*}[tb!]
    \centering
\resizebox{1.5\columnwidth}{!}{
    \begin{tabular}{l|rr|rr|rr|rr|rr}
    \toprule
    &	\multicolumn{2}{c}{\textbf{Fact. A}}	&	\multicolumn{2}{c}{\textbf{Subj. A}}	& \multicolumn{2}{c}{\textbf{Fact. Q}} &	\multicolumn{2}{c}{\textbf{Subj. Q}} & \multicolumn{2}{c}{\textbf{Overall}}		\\
    & \textbf{F1} & \textbf{E} & \textbf{F1} & \textbf{E} & \textbf{F1} & \textbf{E} & \textbf{F1} & \textbf{E} & \textbf{F1} & \textbf{E} \\
    \midrule
Tripadvisor & 17.50 & 20.88 & 1.28 & 7.43 & 18.85 & 21.60 & 1.16 & 7.37    & 1.01 & 7.42    \\
Restaurants & 10.36 & 12.38 & 8.37 & 11.49 & 13.85 & 15.77 & 8.19 & 11.07 & 5.71 & 8.65   \\
Movies      & 14.49 & 14.63 & 5.17 & 8.02 & 14.27 & 14.41 & 5.44 & 8.28  & 3.08 & 5.84    \\
Books       & 13.95 & 14.10 & 7.18 & 9.82 & 14.68 & 14.83 & 7.05 & 9.68   & 4.06 & 6.67    \\
Electronics & 14.15 & 18.70 & 0.28 & 7.06 & 13.29 & 18.22 & 0.40 & 7.24   & -0.01 & 7.26    \\
Grocery     & 9.69  & 11.74 & -0.16 & 3.75 & 10.71 & 12.32 & -0.48 & 3.41 & -1.57 & 2.20    \\
\midrule
Average     & 13.35 & 15.40 & 3.69 & 7.93 & \textbf{14.27} & \textbf{16.19} & 3.63 & 7.84 & 2.05 & 6.34 \\
    \bottomrule
    \end{tabular}
    }
    \caption{MTL gains/losses over the fine-tuning condition (F1 and Exact match), across subj./fact.~QA.}
    \label{tab:subj_aware}
\end{table*}

\subsection{Baselines}

We use four pre-trained models to investigate how their performances on \subjqa{} compare with a factual dataset, SQuAD~\cite{RajpurkarZLL16}, created using Wikipedia. Specifically, we evaluate BiDaF~\citep{bidaf}, FastQA~\citep{fastqa}, JackQA~\citep{jackqa}\footnote{\url{https://github.com/uclnlp/jack}} and BERT~\cite{bert},\footnote{BERT-Large, Cased (Whole Word Masking)} all pre-trained on SQuAD. Additionally, we fine tune the models on each domain in \subjqa{}.

\begin{figure}
    \centering
    \includegraphics[scale=0.35]{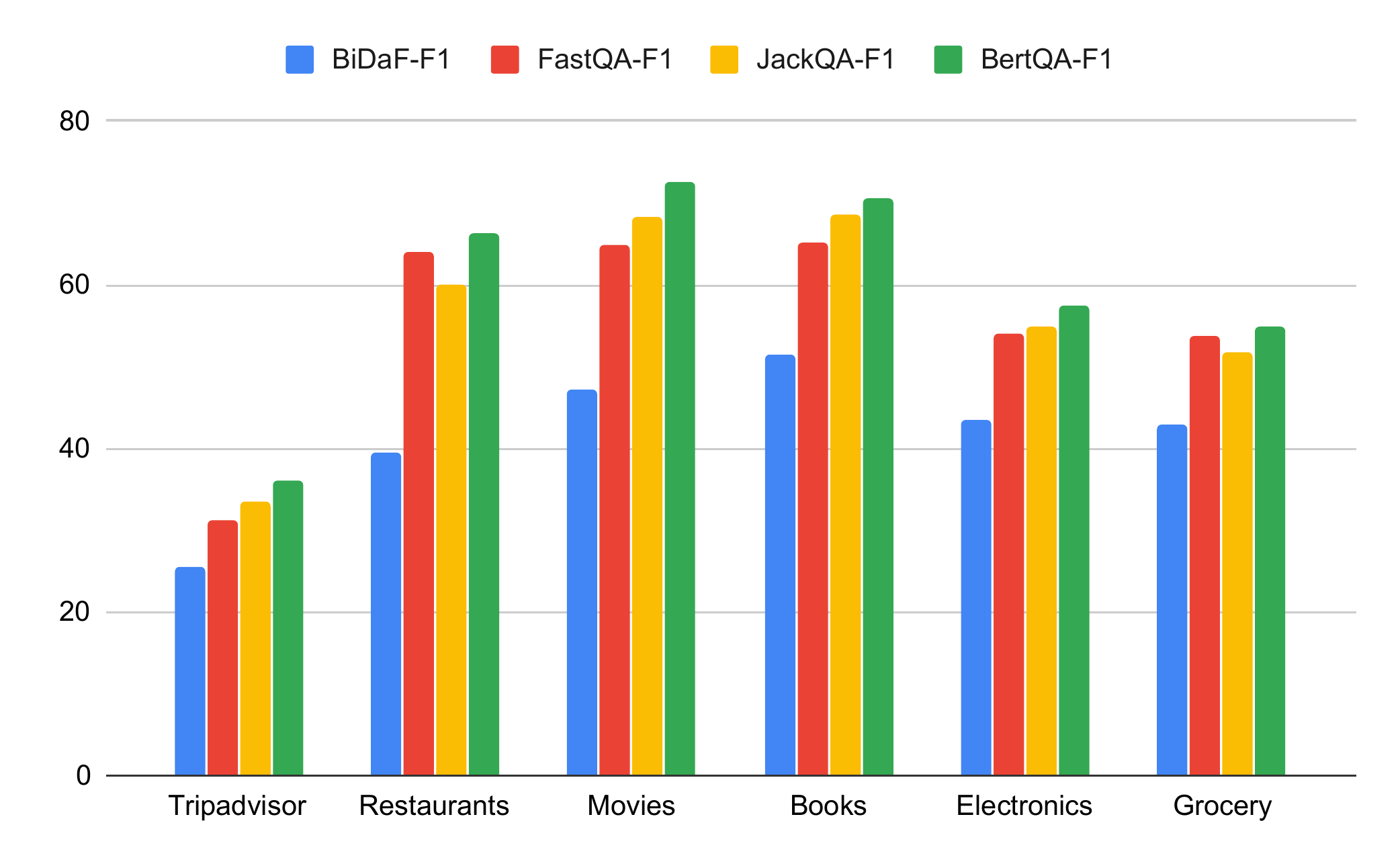}
    \caption{Gain in F1 with models fine-tuned on different domains over the pre-trained model. }
    \label{fig:fine_tuned}
\end{figure}

Figure~\ref{fig:qa_baselines} shows the F1 scores of the pre-trained models. 
We report the Exact match scores in Appendix~\ref{sec:appendix_experiments}. 
Pre-trained models achieve F1 scores as high as 92.9\% on the SQuAD. 
On the other hand, the best model achieves an average F1 of 30.5\% across all domains and 36.5\% F1 at best on any given domain in \subjqa{}.
The difference in performance can be attributed to both differences in domain (Wikipedia vs. customer reviews) and how subjectivity is expressed across different domains.


Figure~\ref{fig:fine_tuned} shows the absolute gains in F1 scores of models fine-tuned on specific domains, over the pre-trained model. After fine-tuning on each domain, the best model achieves an average F1 of 74.1\% across the different domains. 
While fine-tuning significantly boosts the F1 scores in each domain, they are still lower than the F1 scores on the SQuAD dataset.
We argue that this is because the models are agnostic about subjective expressions in questions and reviews. 
To validate our hypothesis, we compare the gain in F1 scores of the BERT model on subjective questions and factual questions. We find that the difference in F1 gains is as high as 23.4\% between factual and subjective questions. F1 gains differ by as much as 23.0\% for factual vs. subjective answers.  

\subsection{Subjectivity-Aware Modeling}
After fine-tuning over each domain in the MTL setting, the subjectivity-aware model achieves an average F1 of 76.3\% across  domains (Table~\ref{tab:subj_aware}). 
Incorporating subjectivity in the model thus boosts performance across all domains and both metrics.
Although there are gains also for subjective questions and answers, it is noteworthy that the highest gains can be found for factual questions and answers.
This is likely because existing techniques already are tuned for factual questions.

\subsection{Error Analysis}
We perform an error analysis based on the predictions of the BERT model fine-tuned for each domain, investigating a sample of 10 questions for which the model's predictions are incorrect.
This reveals that the model often predicts a question to be unanswerable, although a ground truth answer exists.
Interestingly, most of these questions are subjective.
A detailed examination reveals that answers to these questions are relatively indirect, hence it is not surprising that the model struggles.
For instance, given the question \textit{"How slow is the internet service?"}, the correct answer is \textit{"Don't expect to always get the 150Mbps”}.

%% file: 8_related.tex

We are witnessing a substantial rise in user-generated content, including subjective information ranging from personal experiences to opinions about specific aspects of a product.
However, subjectivity has largely been studied in the context of sentiment analysis~\cite{hu2004mining} and opinion mining~\cite{blair2008building}, with a focus on text polarity. 
There is renewed intereste in incorporating subjective opinion data into general data management systems~\cite{li2019subjective,kobren2019constructing} and for querying subjective data. 

In this work, we revisit subjectivity in the context of review QA. 
\citet{yu2012answering,amazonreviews} also use review data, as they leverage question types and aspects to answer questions.
However, 
no prior work has modeled subjectivity explicitly using end-to-end architectures.

Furthermore, none of the existing review-based QA datasets are targeted at understanding subjectivity. This can be attributed to how these datasets are constructed. Large-scale QA datasets, such as SQuAD~\cite{RajpurkarZLL16}, NewsQA~\cite{TrischlerWYHSBS17}, CoQA~\cite{ReddyCM19}, MLQA~\cite{lewis:2020} are based on factual data. We are the first to attempt to create a review-based QA dataset for the purpose of understanding subjectivity.
Recent work has corroborated our findings on the benefits of modelling subjectivity QA, and highlights the differences in the distributions of hidden representation between \subjqa{} and the factual SQuAD data \citep{muttenthaler:2020:unsupervised}.

%% file: appendix.tex
\subsection{Neighborhood Model Construction}

For constructing the matrix for factorization, we focus on frequently reviewed items and frequent extractions. In particular, we consider items which have more than 10,000 reviews and extractions that were expressed in more than 5000 reviews. Once the matrix is constructed, we factorize it using non-negative factorization method using 20 as the dimension of the extraction embedding vector. 

In the next step, we construct the neighborhood model by finding top-10 neighbors for each extraction based on cosine similarity of the extraction and the neighbor. We further select topics from the extractions, and prune the neighbors based on the criteria we described earlier. Table~\ref{tab:kb_examples} shows example extraction and their neighbors discovered using the neighborhood model.

\begin{table}[h]
\centering
\setlength{\tabcolsep}{1pt}
    \begin{tabular}{cll}
    \toprule
    \textbf{Domain} &
    \textbf{Extraction} & \textbf{Neighbor} \\
        \midrule
        \multirow{3}{*}{Movies}
          & \apphrase{(satisfying, ending)} & \apphrase{(good, script)} \\
        & \apphrase{(believable, acting)} & \apphrase{(interesting, movie)}\\
        \midrule
        \multirow{3}{*}{Electronics} 
          & \apphrase{(responsive, key)} & \apphrase{(nice, keyboard)} \\
        & \apphrase{(easy to navigate, menu)} & \apphrase{(simple, interface)}\\
        \midrule
        \multirow{3}{*}{Books}
          & \apphrase{(graphic, violence)} & \apphrase{(disturbing, book)} \\
        & \apphrase{(interesting, twist)} & \apphrase{(unpredictable, story)}\\
        \midrule
        \multirow{3}{*}{Grocery}
          & \apphrase{(healthy, snack)} & \apphrase{(high, protein)} \\
        & \apphrase{(easy to follow, direction)} & \apphrase{(quick, preparation)}\\
        \midrule
        \multirow{3}{*}{Tripadvisor}
          & \apphrase{(excellent, service)} & \apphrase{(amazing, hotel)}\\
        & \apphrase{(good, drinks)} & \apphrase{(great, bar)}\\
        \midrule
        \multirow{3}{*}{Restaurants}
          & \apphrase{(great, meal)} & \apphrase{(good, restaurant)}\\
        & \apphrase{(fast and friendly, service)} & \apphrase{(quick, food)}\\
        \bottomrule
    \end{tabular}
    \caption{Examples extraction and their neighbors}
    \label{tab:kb_examples}
\end{table}

\subsection{Additional Dataset Examples}

Table~\ref{tab:data_examples} shows more examples of question and their review snippets. Answer spans have been underlined. As can be seen, many questions and answer spans in our dataset are subjective.

\begin{table*}
\centering

\setlength{\tabcolsep}{3pt}
    \begin{tabularx}{\textwidth}{c l >{\setlength{\baselineskip}{0.9\baselineskip}}X}
    \toprule
    \textbf{Domain} &
    \textbf{Question} & \textbf{Review} \\
        \midrule
        \multirow{3}{*}{Movies}
          & \apphrase{Is the main character a good actor?} & \apphrase{Beautifully written series depicting the lives of criminals in Kentucky and thedeputies in the US Marshall Service who attempt to stop the criminal elementin the community.The acting is outstanding and the \ul{cast bring the characters to life.}} \\
        & \apphrase{Can we enjoy the movie along with family?} & \apphrase{\ul{An outstanding romantic comedy}, 13 Going on 30, brings to the screen exactly what the title implies: the story of a 13-year old girl who has her wish fulfilled and wakes up seven years later in the body of her 30-year old self!}\\
        \midrule
        \multirow{3}{*}{Electronics} 
          & \apphrase{Why is the camera of poor quality?} & \apphrase{Item like the picture, fast deliver 3 days well packed, good quality for the price. The camera is decent (as phone cameras go), \ul{There is no flash} though.} \\
        & \apphrase{How big is the unit?} & \apphrase{It's a great product, especially for the money.Good Battery life, totally useable for a full movie. Storage capacity.  \ul{32GB} is a great point for the price. Speakers. Screen - bright, clear and HD resolution.}\\
        \midrule
        \multirow{3}{*}{Books}
          & \apphrase{Is the plot line good enough?} & \apphrase{\underline{The book got me hooked almost immediately} and then I got to the end and realized that there is another book after this one. I was unaware of this dilemma but its so good I did not care.  Characters and dialogue are good but I liked the movie better.} \\
        & \apphrase{What is the most exciting part of the story?} & \apphrase{Yann Martel's Life of Pi is a wondrous novel--there is much to wonder and marvel at.  The story is simple, yet complex at the same time and can be read on many levels. \ul{The novel ends with a philisophical bang, which did blow me away.} }\\
        \midrule
        \multirow{3}{*}{Grocery}
          & \apphrase{Does this coffee taste good?} & \apphrase{{While I don't consider myself a coffee snob, I know what I like. Actually, \ul{it tasted on the light side of light}.  I prefer a dark bold flavor but don't mind a good medium roast. The package also said Exotic Floral and Berry.}} \\
        & \apphrase{Is the sauce tasty?} & \apphrase{I was hoping that this sauce would be a little more consistent and thick that it is.  \ul{The taste is a bit sharp and perhaps it's just not in my palette}, but I'll stick with a homeade ranchero until I find one that is a quick retail replacement.}\\
        \midrule
        \multirow{3}{*}{Tripadvisor}
          & \apphrase{How was the attention of the staff?} & \apphrase{The Handlery Union Square Hotel offers great rooms in the centre of San Francisco for a very nice price! Excellent value for money. \ul{Friendly personnel} and top location! Highly recommended!}\\
        & \apphrase{How is the parking?} & \apphrase{\$40.00 a night; not the room - parking rate per night.Got a room with no view at the end of the hall way, nasty smell like very old, damp room. Front desk clark was not friendly, when raised concern about parking rate, he compared that with NYC parking. The \ul{parking rate is ridiculously high}; \$40 a night!!!!!!!!!!! }\\
        \midrule
        \multirow{3}{*}{Restaurants}
          & \apphrase{Is the price economical?} & \apphrase{This place is the best Pho Place in the area. If you are too lazy to drive to Pho Dau Bo then come here for your hangover cure.  \ul{The prices are a little bit on the high side} but that is simply a reflection of the neighbourbood the restaurant is located in.}\\
        & \apphrase{Was it a good place for late night snacking?} & \apphrase{\ul{Lots of different beer choices and liter-ful mugs} that will satiate your European beer craving.  Social seating on long benches (no individual tables) filters out the snobs and the social invalids, so everyone has a great time!}\\
        \bottomrule
    \end{tabularx}
    \caption{Examples questions and review snippets. Answer spans are underlined.}
    \label{tab:data_examples}
\end{table*}

\subsection{Additional Experimental Results}
\label{sec:appendix_experiments}

\begin{figure}[ht]
    \centering
    \includegraphics[scale=0.35]{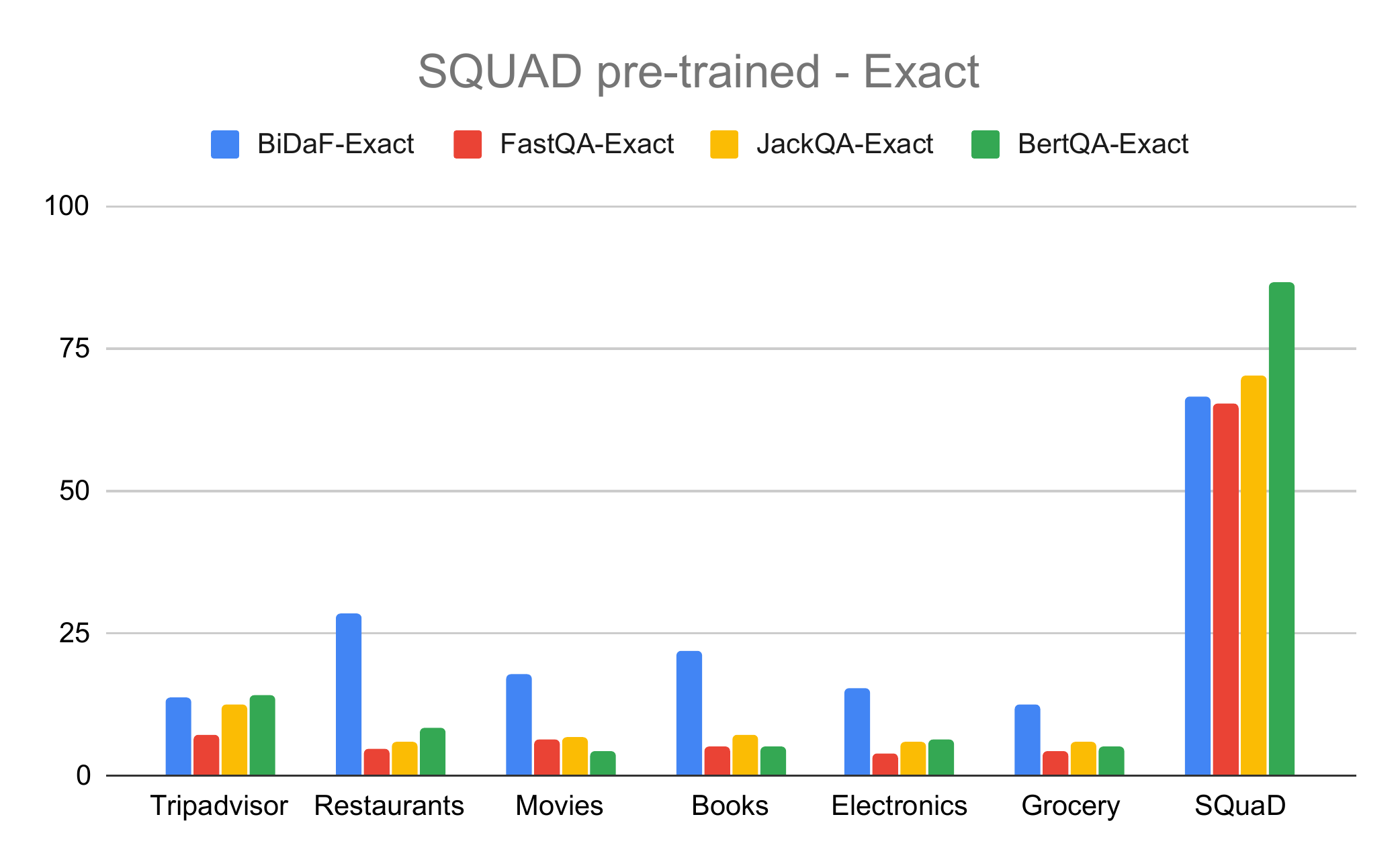}
    \caption{Exact scores of pre-trained out-of-the-box models on different domains. }
    \label{fig:qa_baselines_exact}
\end{figure}

\begin{figure}[ht]
    \centering
    \includegraphics[scale=0.35]{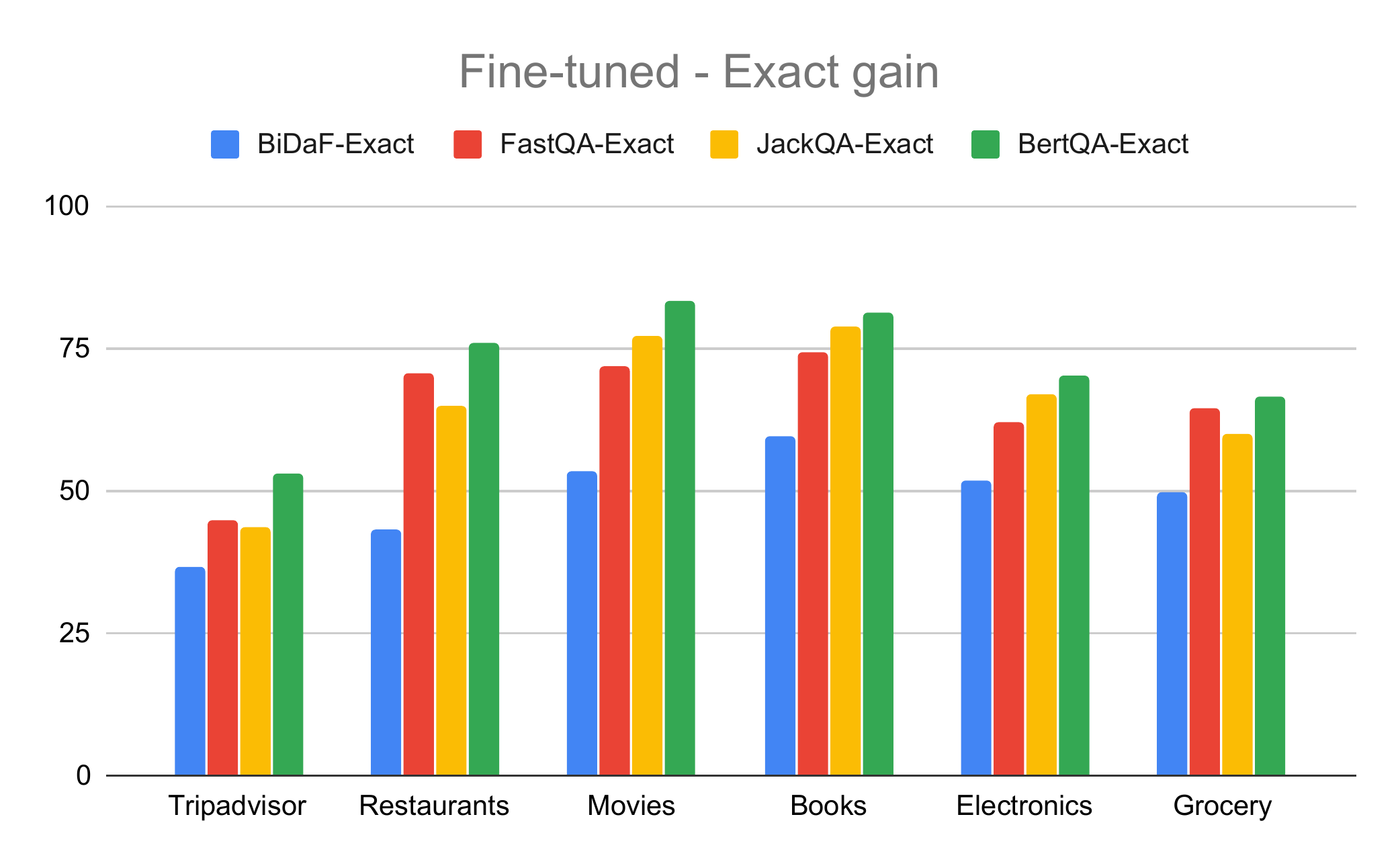}
    \caption{Gain in Exact scores with models fine-tuned on different domains. }
    \label{fig:fine_tuned_exact}
\end{figure}

Figure~\ref{fig:qa_baselines_exact} shows the exact scores achieved by the pretrained out-of-the-box models on various domains in \subjqa{}. Figure~\ref{fig:fine_tuned_exact} shows the exact scores of the models fine-tuned on each domain in \subjqa{}.

\subsection{Crowdsourcing Details}

We used the Appen platform ~\footnote{\url{https://appen.com}} to obtain labels. The platform provides quality control by showing the workers 5 questions at a time, out of which one is labeled by the experts. A worker who fails to maintain 70\% accuracy is kicked out by the platform and his judgements are ignored.

\begin{figure*}
    \centering
    \includegraphics[scale=.5]{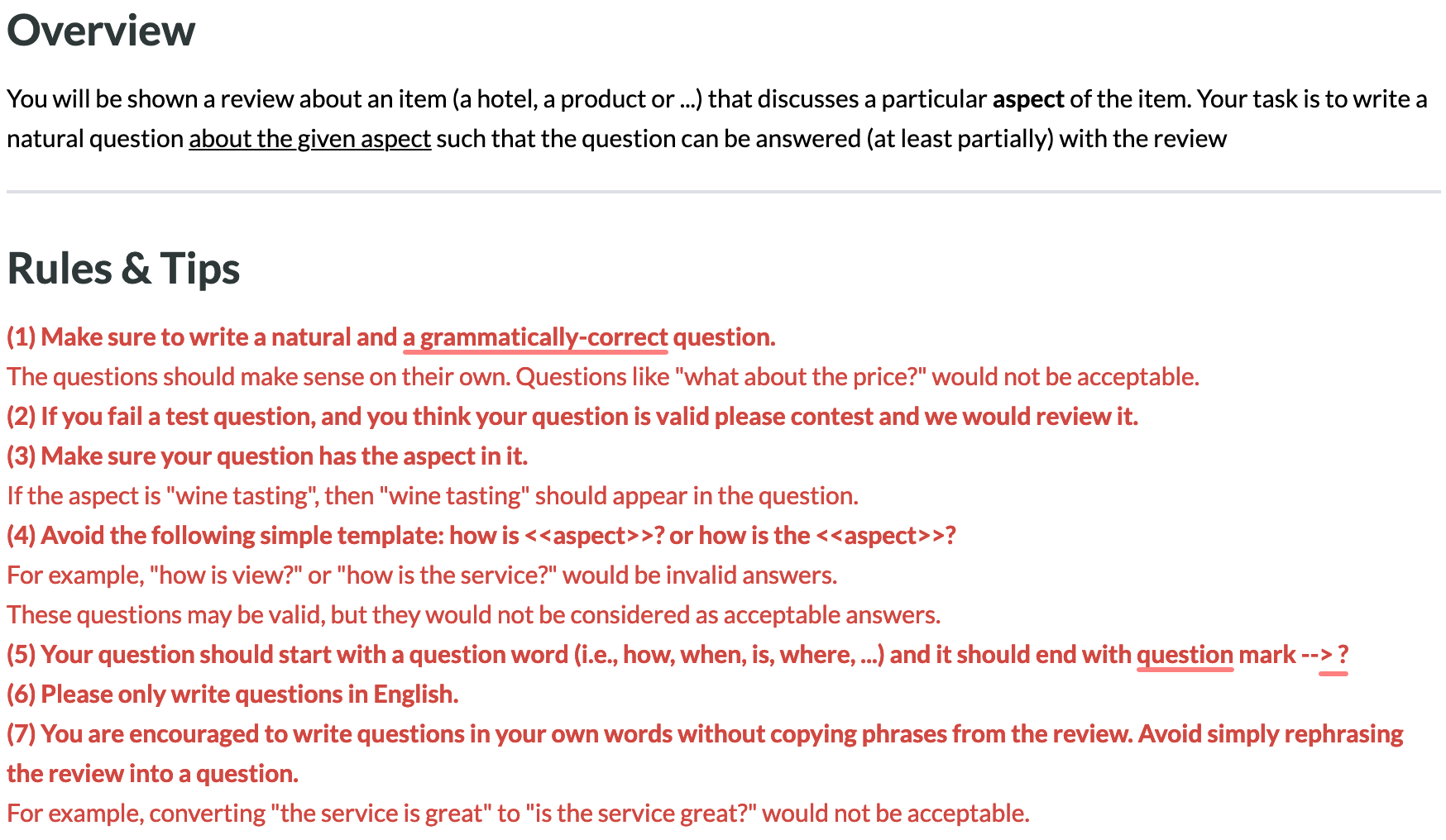}
    \caption{The instructions shown to crowdworkers for the question writing task.}
    \label{fig:questiongeneration}
\end{figure*}

\begin{figure*}
    \centering
    \includegraphics[scale=.55]{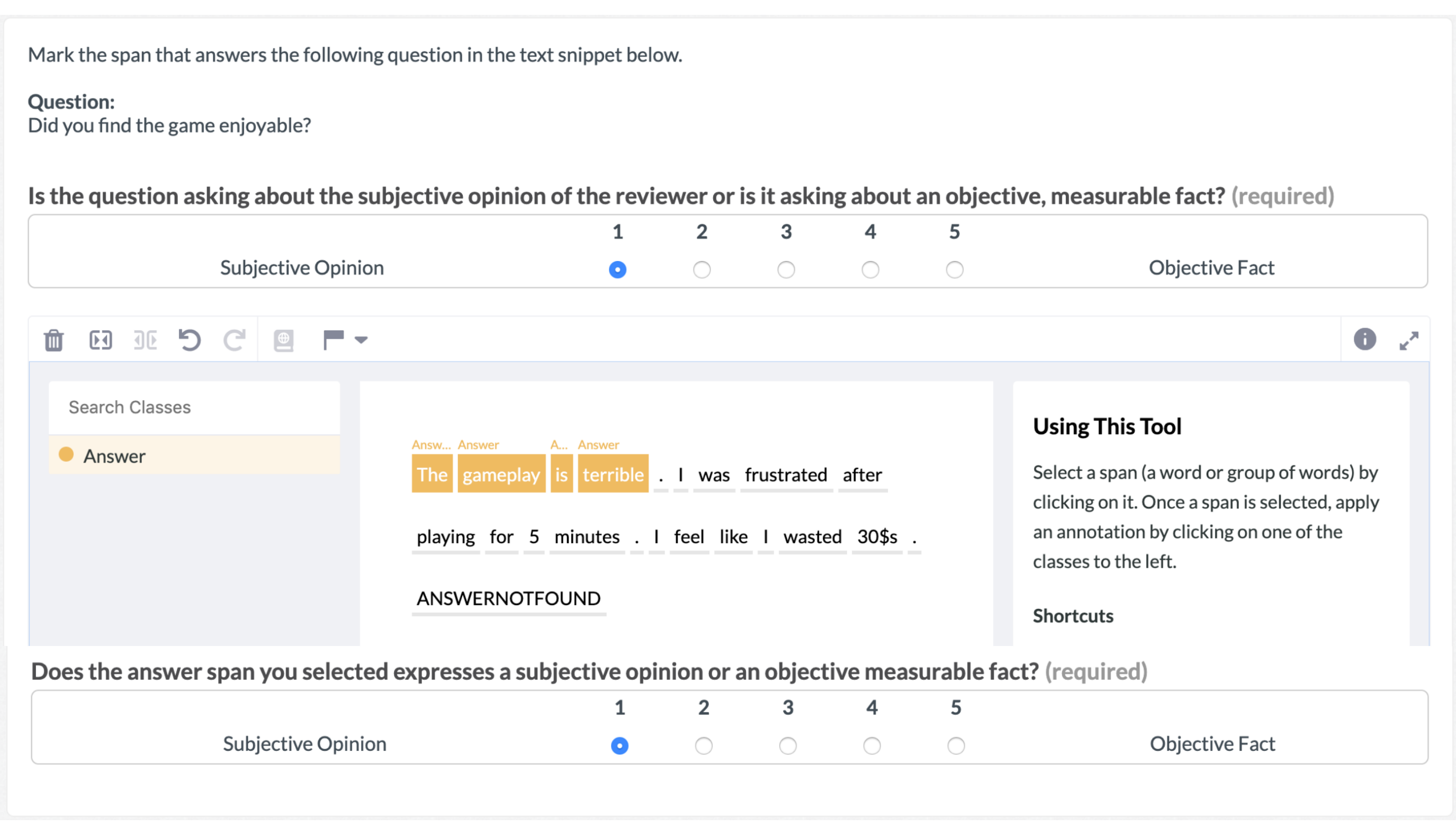}
    \caption{The interface for the answer-span collection and subjectivity labeling tasks.}
    \label{fig:spanselection}
\end{figure*}

Figure~\ref{fig:questiongeneration} illustrates the instructions that were shown to the crowdworkers for the question generation task. Figure~\ref{fig:spanselection} shows the interface for the answer-span collection and subjectivity labeling tasks. The workers assign subjectivity scores (1-5) to each question and the selected answer span. They can also indicate if a question cannot be answered from the given review. Before finalizing this interface, we experimented with multiple design variations on a small subset. This included collecting binary labels for subjectivity instead of scores, simultaneously collecting different labels vs in different tasks etc. To ensure good quality labels, we each worker 5 cents per annotation.

\label{sec:questiongeneration}

\label{sec:spanselection}
